\PassOptionsToPackage{pagenumbers}{wacv}
\documentclass[10pt,twocolumn,letterpaper]{article}
\usepackage[accsupp]{axessibility}
\usepackage{wacv}
\usepackage{multirow}
\usepackage{microtype}
\usepackage{algorithm}
\usepackage{algpseudocode}
\usepackage[most]{tcolorbox}

\makeatletter
\providecommand*{\theHALG@line}{}
\renewcommand*{\theHALG@line}{\thealgorithm.\arabic{ALG@line}}
\makeatother

\newtcolorbox{promptbox}[1][]{
  colback=blue!5!white,
  colframe=blue!75!black,
  fonttitle=\bfseries,
  title=#1,
  arc=1mm,
  boxrule=0.5pt,
  left=2mm,
  right=2mm,
  top=2mm,
  bottom=2mm,
  breakable,
  enhanced,
  width=\linewidth,
  fontupper=\small\ttfamily
}

\definecolor{wacvblue}{rgb}{0.21,0.49,0.74}
\usepackage[breaklinks,colorlinks,allcolors=wacvblue]{hyperref}

\newcommand{\codeurl}{https://github.com/819346675/Think-Capture}
\newcommand{\codeavailability}{\space Our code is available at \href{\codeurl}{\nolinkurl{github.com/819346675/Think-Capture}}.}

\def\wacvPaperID{560}
\def\confName{WACV}
\def\confYear{2027}

\title{Think-Capture: Active 3D Geometry Acquisition for Spatial Reasoning in Vision-Language Models}
\author{Junjie Wang\textsuperscript{1,2}, Zhihao Yuan\textsuperscript{1,2}, Yunhan Tang\textsuperscript{3}, Sheng Xu\textsuperscript{2}\\
Jiaqi Mao\textsuperscript{1,2}, Xiaoyu Ye\textsuperscript{1,2}, Yueru Luo\textsuperscript{1,2}, Xu Yan\textsuperscript{1,2}\\
Chun-Mei Feng\textsuperscript{4}, Zhen Li\textsuperscript{2,1,*}\\[2pt]
{\large \textsuperscript{1}FNii-Shenzhen \quad \textsuperscript{2}CUHK-Shenzhen}\\
{\large \textsuperscript{3}University of Electronic Science and Technology of China}\\
{\large \textsuperscript{4}University College Dublin}\\
{\normalsize \textsuperscript{*}Corresponding author: \texttt{lizhen@cuhk.edu.cn}}}

\hypersetup{pdftitle={Think-Capture: Active 3D Geometry Acquisition for Spatial Reasoning in Vision-Language Models},pdfauthor={Junjie Wang, Zhihao Yuan, Yunhan Tang, Sheng Xu, Jiaqi Mao, Xiaoyu Ye, Yueru Luo, Xu Yan, Chun-Mei Feng, Zhen Li}}

\usepackage{placeins}
\hypersetup{pdfsubject={Main paper, references, and supplementary material}}
\begin{document}
\maketitle
\begin{abstract}
Vision-Language Models (VLMs) excel at 2D recognition but often fail to maintain consistent 3D relations across sparse viewpoints, leading to spatial hallucinations under occlusion and perspective changes. Existing methods commonly fuse all frames with dense geometry, incurring substantial computation and context overhead while encouraging blind reliance on imperfect estimators. We present \textbf{Think-Capture}, a query-conditioned framework that first identifies task-critical objects and then acquires their geometry on demand. A 3D toolkit lifts zero-shot masks into a common coordinate frame, and a Symbolic 3D State Abstraction compresses object and camera states into structured text. Hierarchical Group Relative Policy Optimization (GRPO) trains the VLM on cases in which geometry either helps or misleads, teaching it when to use tool evidence and when to fall back on visual cues. Think-Capture-4B reaches 54.4\% on MindCube-Tiny (+21.5 points over its backbone), 39.1\% on the MMSI-Bench positional subset (+7.9), and 42.8\% on ViewSpatial (+5.2). It also remains robust under coordinate and mask perturbations. The full stack fits one 24\,GB GPU and serves as an asynchronous high-level reasoner on an Agibot~G1.\codeavailability

\end{abstract}

\section{Introduction}
\label{sec:intro}

Multi-view spatial reasoning is essential for embodied systems that must navigate and manipulate objects~\cite{ji2025robobrain,yang2025embodiedbench,ziliotto2025tango,intelligence2025pi,black2024pi_0,intelligence2025pi_,xu2026reaction}. Although recent VLMs are strong at image understanding~\cite{zhu2025internvl3,bai2025qwen3,team2025kimi}, they struggle with cross-view consistency, occlusion, and relative 3D position~\cite{deng20253d,wuspatial,li2025spatialladder,yang2025thinking,yin2025spatial,yuan2025scene,yang2025mmsi,li2025viewspatial,yu2025far}. Under depth or viewpoint ambiguity, they often substitute image-plane heuristics for a persistent spatial model~\cite{yin2025spatial}.

\begin{figure}[t]
    \centering
    \includegraphics[width=\linewidth]{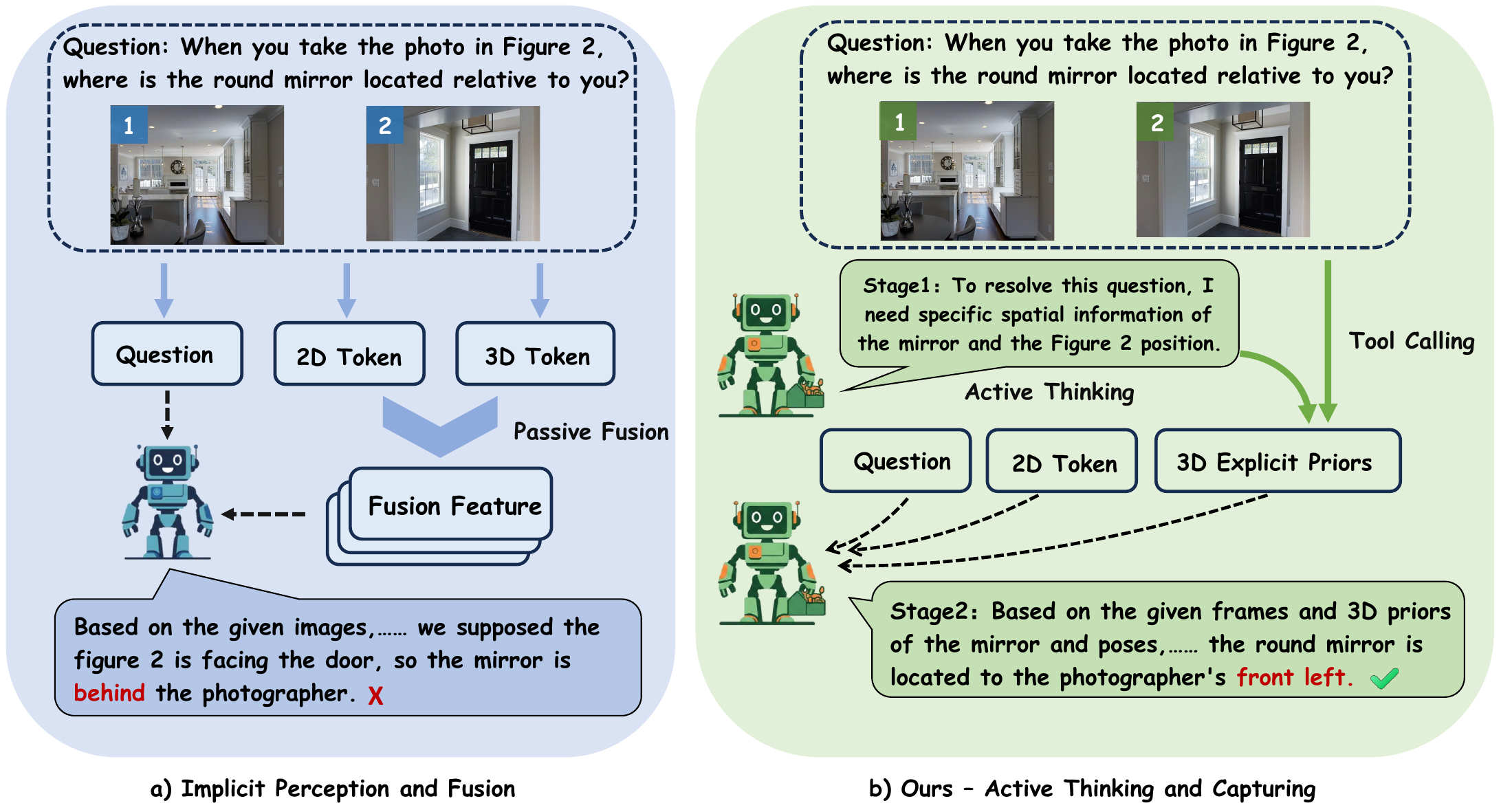}
    \caption{\textbf{Spatial reasoning paradigms.} \textbf{(a)} Passive methods fuse 2D and 3D features for every query. \textbf{(b)} Think-Capture first selects task-critical targets, acquires their geometry, and reasons jointly over visual tokens and explicit 3D state.}
    \label{fig:method_comparison}
\end{figure}

Most existing approaches follow an \textit{implicit perception-and-fusion} paradigm (Fig.~\ref{fig:method_comparison}a): they encode dense scene geometry and fuse it with image features before seeing the query. Processing task-irrelevant geometry increases computation and context length and can introduce distracting evidence~\cite{yin2025spatial,fan2025vlm,maspatialreasoner}. Agentic alternatives let VLMs invoke external 3D tools~\cite{zhang2026think3d}. These interfaces make geometric evidence explicit, but errors in the tool output or its interpretation can still propagate to the answer.

Think-Capture instead performs query-conditioned acquisition (Fig.~\ref{fig:method_comparison}b). The VLM names only the objects needed to answer the question; a 3D Capturing Toolkit then estimates their masks, coordinates, and cross-view relations. A \textit{Symbolic 3D State Abstraction} converts these measurements into compact object and camera descriptions, filtering irrelevant geometry while preserving the relations required by the query~\cite{carion2025sam,wang2025pi}.

We train a 4B backbone with GRPO~\cite{guo2025deepseek,ouyang2025spacer,shao2024deepseekmath}. The training set distinguishes queries for which 3D evidence corrects the vision-only model from those for which imperfect geometry is misleading. Hierarchical rewards encourage correct use of helpful geometry and incorporate negative examples to retain correct answers when tool augmentation is misleading. The resulting policy learns to combine explicit 3D state with the original frames instead of trusting either source unconditionally.

We evaluate on MindCube~\cite{yin2025spatial}, MMSI-Bench~\cite{yang2025mmsi}, and ViewSpatial-Bench~\cite{li2025viewspatial}, covering mental rotation, cross-view relations, and perspective taking. We further test backbone scaling, zero-shot transfer, controlled tool corruption, and a matched 2D-box baseline. Finally, we deploy Think-Capture as an asynchronous high-level spatial reasoner on a real Agibot~G1.

In summary, our main contributions are:
\begin{itemize}
    \item We introduce query-conditioned 3D acquisition and a compact symbolic interface that exposes task-relevant object and camera geometry to a VLM.
    \item We propose hierarchical 2D--3D GRPO rewards that promote useful geometry integration and improve robustness to noisy tool evidence.
    \item We demonstrate consistent gains across backbones and benchmarks, robustness to controlled tool corruption, and real-robot integration on consumer hardware.
\end{itemize}

\section{Related Work}

\subsection{Implicit Spatial Learning in VLMs}
Object-centric grounding~\cite{yuan2021instancerefer,liu2024grounding} and 3D-informed learning~\cite{ma2025spatialllm,wuspatial,li2025spatialladder,xu2025multi,yang2025cambrian,chen2025think} enrich spatial representations. Visual-tool use and video reasoning provide complementary strategies~\cite{su2025openthinkimg,su2025thinking,feng2025video}. Geometry estimators such as VGGT and MapAnything recover scene structure from images~\cite{wang2025vggt,keetha2025mapanything}. These advances motivate combining learned visual priors with explicit, query-relevant geometric evidence.

\subsection{Spatial Reasoning From Multi-views}
Multi-view benchmarks test cross-view understanding~\cite{yeh2025seeing}, mental rotation~\cite{yin2025spatial}, relations between entities that never co-occur~\cite{yang2025mmsi}, and perspective transformations~\cite{li2025viewspatial}. Mental models provide a cognitive account of internal spatial representations~\cite{johnson1980mental,johnson1983mental}; recent evaluations reveal gaps in VLM spatial understanding~\cite{daxberger2025mm,ramakrishnan2024does}. Perspective-aware reasoning and spatial-reasoning surveys motivate structured geometry as a complement to visual cues~\cite{lee2025perspective,zha2025enable}.

\subsection{Explicit Spatial Reasoning}
MindCube~\cite{yin2025spatial} trains VLMs to generate cognitive maps, whereas GPT4Scene-style systems~\cite{qi2025gpt4scene,zhenglearning} provide externally constructed metric or bird's-eye-view maps. Both reconstruct-then-reason approaches represent the whole scene. Tool-using methods such as Think3D~\cite{zhang2026think3d} instead let an agent manipulate point clouds or estimate camera anchors~\cite{wang2025pi}. Think-Capture differs in three coupled design choices: query-conditioned target selection, a common symbolic interface for object and camera state, and hierarchical GRPO that trains the VLM to accept or reject imperfect tool evidence. The controlled corruption and matched 2D-box experiments in Sec.~\ref{sec:experiment} evaluate robustness and the value of the symbolic 3D representation.

\section{Method}
\label{sec:method}

\subsection{Overview: The Think-Capture Paradigm}
\label{subsec:overview}
Think-Capture implements a two-stage Reason-then-Act loop: it first plans which geometry is needed, then reasons over the images and the captured symbolic state.

\begin{figure*}[t]
  \centering
  \includegraphics[width=\linewidth]{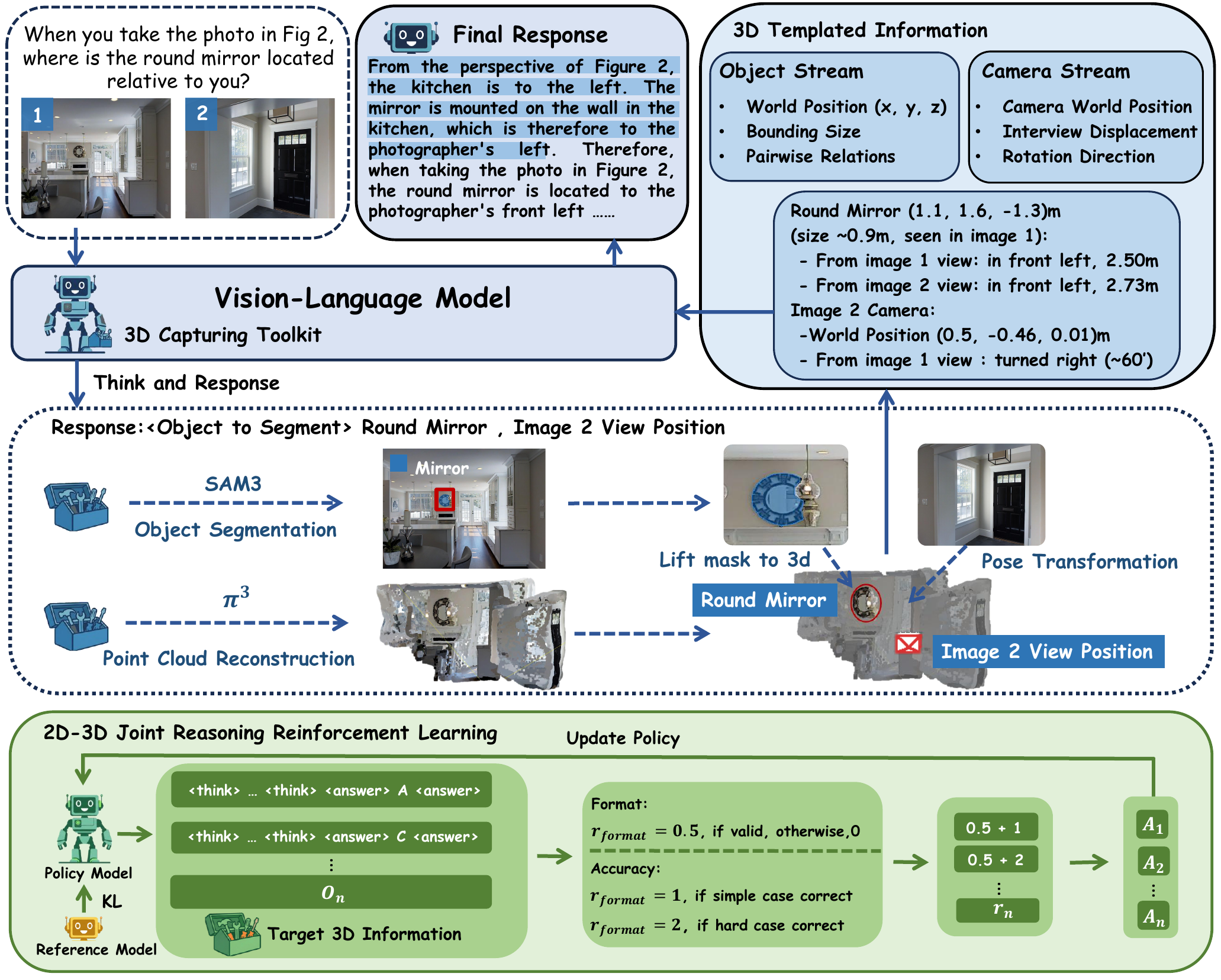}
  \caption{\textbf{Think-Capture overview.} Stage~1 selects query-relevant targets. SAM3 and $\pi^3$ produce the object and camera state used with the images in Stage~2. In the training panel, the accuracy-branch labels denote $r_{\mathrm{acc}}$, with $+1/+2$ for correct positive/negative examples (Sec.~\ref{subsec:grpo}). The example annotations illustrate the interface.}
  \label{fig:method_framework}
\end{figure*}

\textbf{Problem formulation.} Let $\mathcal{V}=\{v_k\}_{k=0}^{K-1}$ be $K$ views of a scene and $\mathcal{Q}$ a spatial query. A VLM policy $\pi_\theta$ generates a reasoning trajectory $y=(y^1,\ldots,y^L)$ and final answer $\mathcal{A}$ conditioned on $(\mathcal{V},\mathcal{Q})$.

Given $(\mathcal{V},\mathcal{Q})$, Stage~1 emits semantic targets inside an \texttt{<objects\_to\_segment>} block. The toolkit acquires only the requested object geometry and returns a symbolic text state. Stage~2 then combines that state with the original images to produce $\mathcal{A}$ (Fig.~\ref{fig:method_framework}).

\subsection{3D Perception via Tool-Calling}
\label{subsec:active_perception}
For each target $t$, SAM3~\cite{carion2025sam} produces a zero-shot mask $M^{(k)}\in\{0,1\}^{H\times W}$ in each relevant frame.

In parallel, the feed-forward estimator $\pi^3$ (Pi3X, with approximate metric scale) predicts a local point map $P_{\mathrm{local}}^{(k)}\in\mathbb{R}^{H\times W\times3}$ and camera-to-world pose $T_{\mathrm{cam}}^{(k)}=[R^{(k)}\mid t^{(k)}]\in SE(3)$.

We use the first frame as the world origin, so $T_{\mathrm{cam}}^{(0)}=I_{4\times4}$. A local point $p_{\mathrm{local}}$ is transformed as
\begin{equation}
    \begin{bmatrix} p_{\mathrm{world}} \\ 1 \end{bmatrix} = \begin{bmatrix} R^{(k)} & t^{(k)} \\ \mathbf{0}^T & 1 \end{bmatrix} \begin{bmatrix} p_{\mathrm{local}} \\ 1 \end{bmatrix}.
\end{equation}

The mask selects the corresponding world-space points:
\begin{equation}
P_{\mathrm{world}}^{(k)}=\left\{R^{(k)}P_{\mathrm{local}}^{(k)}[i,j]+t^{(k)}\;\middle|\;M^{(k)}[i,j]=1\right\}.
\end{equation}

Only masked target points are aggregated into object state; the system neither serializes nor sends the dense scene to the VLM. Supplementary Sec.~B gives the complete pipeline and four-frame budget.

\textbf{Role and computational benefit of target selection.} Stage~1 selects semantic object targets conditioned on the question, rather than requesting every object in the scene. For $K$ retained frames and $m$ targets, the target-dependent work comprises text-prompted segmentation, masked-point aggregation, and up to $O(m^2)$ object-pair relations. Restricting $m$ therefore avoids processing and describing irrelevant objects. The four-frame cap bounds dense $\pi^3$ estimation, while target selection controls segmentation and symbolic processing within those frames. Our matched 2D-box control compares Stage~2 representations using the same selected targets and masks.

\subsection{3D Templated Information Generation}
\label{subsec:templated_feedback}

Passing dense point clouds to a VLM creates a modality gap and consumes substantial context. We instead distill the acquired geometry into two structured text streams (Fig.~\ref{fig:method_framework}):

\begin{itemize}
    \item \textbf{Object stream ($\mathcal{O}$).} Each target $i$ is represented by world position $p_i$, bounding-box scale $s_i$, and pairwise relations $\mathcal{R}_{i,j}$. We compute
    \begin{equation}
        d_{i,j} = \| p_i - p_j \|_2, \quad \vec{v}_{i,j} = \frac{p_i - p_j}{\| p_i - p_j \|_2}.
    \end{equation}
    and map them to templates such as ``front-left, distance $d_{i,j}$.''
    
    \item \textbf{Camera stream ($\mathcal{C}$).} Each frame stores $(T_{\mathrm{cam}}^{(k)},\Delta T^{(k,k-1)},\vec d_k)$, where
    \begin{equation}
        \Delta T^{(k, k-1)} = \left( T_{cam}^{(k-1)} \right)^{-1} T_{cam}^{(k)}.
    \end{equation}
    is decomposed into translation and Euler angles, yielding descriptions such as ``turned left $\sim76^\circ$.''
\end{itemize}

This bottleneck filters task-irrelevant geometry and uses 100 tokens per query on average, versus 205 for the unstructured tool output (a 51\% reduction). Back-projection and formatting details appear in supplementary Sec.~B.
\begin{table*}[t]
\centering
\caption{\textbf{Comprehensive performance comparison} on MindCube-Tiny, MMSI-Bench (Positional Relationship Subset), and ViewSpatial-Bench. Metrics are reported as accuracy (\%). We grey out the proprietary models. The best performance among open-source models is highlighted in \textbf{bold}. The improvement of \textbf{Think-Capture} over the baseline (\textit{Qwen3-VL-4B}) is shown in \textcolor{wacvblue}{blue} below the score.}
\label{tab:main_results}
\resizebox{\textwidth}{!}{
\begin{tabular}{lcccccccccccc}
\toprule
\multirow{2}{*}{\textbf{Model}} & \multicolumn{4}{c|}{\textbf{MindCube-Tiny}} & \multicolumn{7}{c|}{\textbf{MMSI-Bench-Positional Relationship Subset}} & \textbf{ViewSpatial} \\
\cmidrule(lr){2-5} \cmidrule(lr){6-12} \cmidrule(lr){13-13}
 & Among & Around & Rotation & \textbf{Overall} & Cam-Cam & Cam-Obj & Obj-Reg & Cam-Reg & Reg-Reg & Obj-Obj & \textbf{Overall} & \textbf{Overall} \\
\midrule
\multicolumn{13}{l}{\textit{\textcolor{gray}{Proprietary Models}}} \\
\textcolor{gray}{GPT-4o} & \textcolor{gray}{44.6} & \textcolor{gray}{58.4} & \textcolor{gray}{39.0} & \textcolor{gray}{46.8} & \textcolor{gray}{34.4} & \textcolor{gray}{19.8} & \textcolor{gray}{37.6} & \textcolor{gray}{27.7} & \textcolor{gray}{23.5} & \textcolor{gray}{24.5} & \textcolor{gray}{28.0} & \textcolor{gray}{35.0} \\
\textcolor{gray}{Gemini-2.5 Pro} & \textcolor{gray}{40.2} & \textcolor{gray}{56.7} & \textcolor{gray}{84.3} & \textcolor{gray}{52.5} & \textcolor{gray}{39.7} & \textcolor{gray}{45.3} & \textcolor{gray}{35.2} & \textcolor{gray}{43.3} & \textcolor{gray}{39.5} & \textcolor{gray}{31.9} & \textcolor{gray}{39.0} & \textcolor{gray}{43.6} \\
\textcolor{gray}{Doubao-1.5} & \textcolor{gray}{46.0} & \textcolor{gray}{38.4} & \textcolor{gray}{75.1} & \textcolor{gray}{49.7} & \textcolor{gray}{25.8} & \textcolor{gray}{33.7} & \textcolor{gray}{49.4} & \textcolor{gray}{39.8} & \textcolor{gray}{33.3} & \textcolor{gray}{33.0} & \textcolor{gray}{35.6} & - \\
\textcolor{gray}{GLM-4.5V} & \textcolor{gray}{44.1} & \textcolor{gray}{23.5} & \textcolor{gray}{46.9} & \textcolor{gray}{39.7} & \textcolor{gray}{38.7} & \textcolor{gray}{30.2} & \textcolor{gray}{40.0} & \textcolor{gray}{39.8} & \textcolor{gray}{29.6} & \textcolor{gray}{35.1} & \textcolor{gray}{35.6} & - \\
\midrule
\multicolumn{13}{l}{\textit{Open-Source Models}} \\
InternVL3-8B & 39.2 & \textbf{62.1} & 34.0 & 43.7 & 25.8 & 25.6 & 35.3 & 28.9 & \textbf{37.0} & 31.9 & 30.6 & 38.6 \\
InternVL3-38B & 48.5 & 56.0 & 32.5 & 47.2 & 18.3 & 22.1 & \textbf{38.8} & 34.9 & 35.8 & 22.3 & 28.3 & 41.2 \\
Qwen2.5-VL-7B & 32.1 & 37.2 & 38.8 & 34.6 & 24.7 & 25.6 & 29.4 & 26.5 & 24.7 & 24.5 & 25.9 & 36.8 \\
Qwen3-VL-4B (Base) & 26.5 & 44.6 & 37.5 & 32.9 & 28.0 & 41.9 & 28.2 & 31.3 & 29.6 & 28.7 & 31.2 & 37.6 \\
\midrule
\multicolumn{13}{l}{\textit{Spatial-LLMs}} \\
Spatial-MLLM-3B & 27.6 & 35.2 & 33.0 & 30.4 & 23.6 & 34.9 & 32.9 & 24.1 & \textbf{37.0} & 26.6 & 29.7 & 36.9 \\
SpatialLadder-3B & 47.1 & 40.2 & 31.5 & 42.5 & 26.3 & 36.4 & 31.2 & 26.0 & 36.2 & 29.9 & 30.9 & 39.8 \\
\midrule
\multicolumn{13}{l}{\textit{Tool-Augmented Method}} \\
Think3D-4B & 39.2 & 35.0 & 39.2 & 38.2 & - & - & - & - & - & - & - & - \\
{\textbf{Think-Capture-4B}} & \textbf{57.6} & 54.1 & \textbf{45.0} & \textbf{54.4} & \textbf{34.5} & \textbf{50.1} & 33.2 & \textbf{50.3} & 29.9 & \textbf{36.9} & \textbf{39.1} & \textbf{42.8} \\
 & \scriptsize{\textcolor{wacvblue}{(+31.1)}} & \scriptsize{\textcolor{wacvblue}{(+9.5)}} & \scriptsize{\textcolor{wacvblue}{(+7.5)}} & \scriptsize{\textcolor{wacvblue}{(+21.5)}} & \scriptsize{\textcolor{wacvblue}{(+6.5)}} & \scriptsize{\textcolor{wacvblue}{(+8.2)}} & \scriptsize{\textcolor{wacvblue}{(+5.0)}} & \scriptsize{\textcolor{wacvblue}{(+19.0)}} & \scriptsize{\textcolor{wacvblue}{(+0.3)}} & \scriptsize{\textcolor{wacvblue}{(+8.2)}} & \scriptsize{\textcolor{wacvblue}{(+7.9)}} & \scriptsize{\textcolor{wacvblue}{(+5.2)}} \\
\bottomrule
\end{tabular}
}
\end{table*}
\subsection{2D-3D Joint Reasoning}
\label{subsec:grpo}
We fine-tune Qwen3-VL-4B with GRPO~\cite{guo2025deepseek} so that it can combine the target geometry with visual evidence without blindly trusting the tool.

GRPO avoids a separate critic by estimating the baseline from $G$ sampled trajectories. For brevity, writing the clipped objective at the trajectory level gives
\begin{equation}
\begin{aligned}
    \mathcal{J}_{\mathrm{GRPO}}(\theta)
    &= \mathbb{E}_{\mathcal{Q},\mathcal{V}\sim\mathcal{D}}
       \left[\frac{1}{G}\sum_{i=1}^{G}
       \left(\mathcal{L}_i-\beta\mathbb{D}_{\mathrm{KL}}
       (\pi_\theta\|\pi_{\mathrm{ref}})\right)\right], \\
    \mathcal{L}_i
    &= \min\!\left(\rho_i(\theta)A_i,
       \bar{\rho}_i(\theta)A_i\right), \\
    \bar{\rho}_i(\theta)
    &= \operatorname{clip}\!\left(\rho_i(\theta),1-\epsilon,1+\epsilon\right).
\end{aligned}
\end{equation}
where $\rho_i(\theta)=\pi_\theta(y_i\mid\mathcal{Q},\mathcal{V})/\pi_{\theta_{\mathrm{old}}}(y_i\mid\mathcal{Q},\mathcal{V})$, $\bar\rho_i$ is clipped to $[1-\epsilon,1+\epsilon]$, and $\beta$ weights the KL penalty to $\pi_{\mathrm{ref}}$. For groups with nonzero reward variance, the normalized advantage is
\begin{equation}
    A_i = \frac{R(y_i) - \mu_R}{\sigma_R},
\end{equation}
where $\mu_R$ and $\sigma_R$ are the group reward mean and standard deviation. A constant-reward group contains no within-group preference signal.

The reward combines answer accuracy and format compliance:
\begin{equation}
    R(y_i) = r_{acc}(y_i) + r_{format}(y_i).
\end{equation}
We set $r_{\mathrm{format}}=0.5$ when the output follows the required \texttt{<think>...</think><answer>...</answer>} structure.

We run the base model on 10,000 MindCube training samples in vision-only ($\mathcal{M}_{\mathrm{base}}$) and fixed-prompt tool-augmented ($\mathcal{M}_{\mathrm{tool}}$) modes. Their disagreement yields 3,032 training examples in two subsets:
\begin{itemize}
    \item \textbf{Positive samples:} $\mathcal{M}_{\mathrm{base}}$ fails and $\mathcal{M}_{\mathrm{tool}}$ succeeds, so useful geometry should be exploited. Correct answers receive $r_{\mathrm{acc}}=1$.
    \item \textbf{Negative samples:} $\mathcal{M}_{\mathrm{base}}$ succeeds and $\mathcal{M}_{\mathrm{tool}}$ fails, indicating misleading geometry or failed integration. Correct answers despite this interference receive $r_{\mathrm{acc}}=2$.
\end{itemize}

\textbf{Meaning of the hierarchical reward.} The hierarchy is sample-conditioned: the positive subset contains 1,927 examples and the negative subset contains 1,105 examples, determined by the two screening runs above. Cases in which the screening models agree are not retained. These labels describe screening-time disagreement; question difficulty can vary within either subset. The negative subset captures observed regressions under tool augmentation, which can arise from erroneous geometry or incorrect integration of otherwise useful evidence. Its purpose is to train against such regressions while the positive subset promotes useful tool integration. The $+1/+2$ values specify sample-conditioned raw outcome rewards, and the optimization signal is determined by the total rewards after within-group normalization. Supplementary Sec.~C.1 compares this reward design with a variant using no distinction between positive and negative outcome rewards.

\section{Experiment}
\label{sec:experiment}

\subsection{Datasets}
We evaluate Think-Capture on three challenging benchmarks designed to probe 3D spatial reasoning capabilities:
\begin{itemize}
    \item \textbf{MindCube-Tiny}~\cite{yin2025spatial} contains 1,050 questions: 600 Among, 250 Around, and 200 Rotation questions. Accuracy is computed over the valid evaluated examples in each run after excluding rare invalid evaluation cases; the category counts are nominal split sizes, not necessarily the per-run denominators.
    \item \textbf{MMSI-Bench}~\cite{yang2025mmsi} requires relating entities across disjoint frames. The main comparison uses its positional-relation subset; the supplement also reports the full benchmark.
    \item \textbf{ViewSpatial-Bench}~\cite{li2025viewspatial} evaluates perspective taking across egocentric and allocentric reference frames.
\end{itemize}

\subsection{Implementation Details}
We train Qwen3-VL-4B-Instruct on the 3,032 disagreement-mined MindCube training examples, with learning rate $5\times10^{-7}$ and $G=8$ trajectories per prompt. The vision encoder is frozen, and the effective batch size is 16 with gradient accumulation. Training uses four A100 (80\,GB) GPUs for approximately 38 hours, stopping at 1,600 training steps (approximately 152 GPU-hours for this training run). Correct positive and negative samples receive $+1$ and $+2$, respectively, and format compliance receives $+0.5$. Supplementary Sec.~B provides the optimizer, clipping and KL settings, prompts, and resource details.

\textbf{Additional training configuration.} We use AdamW with a cosine learning-rate schedule and 50 warmup steps. The GRPO clipping parameter is 0.2, and the KL coefficient is 0.04. Training uses BF16 and DeepSpeed ZeRO-2 with the vision encoder frozen, reaching a peak GPU memory usage of 75.6\,GB. The reported 152 GPU-hours refer to the 1,600-step GRPO training run. At inference, the toolkit retains at most four frames, while query-conditioned target selection restricts segmentation and symbolic aggregation.

\subsection{Main Results}
Table~\ref{tab:main_results} compares Think-Capture with proprietary VLMs, open-source VLMs, spatially specialized models, and the tool-augmented Think3D-4B~\cite{zhang2026think3d,comanici2025gemini,hurst2024gpt,zeng2025glm}.

\textbf{MindCube-Tiny.} Think-Capture obtains 54.4\%, an absolute gain of \textbf{21.5 points} over Qwen3-VL-4B and a result comparable to much larger proprietary models. The largest gain is on ``Among'' (+31.1), where explicit target geometry helps localize objects in spatial clusters.

\textbf{Unseen benchmarks and backbones.} Without training on either benchmark, Think-Capture improves the MMSI positional subset from 31.2 to 39.1\% (+7.9) and ViewSpatial from 37.6 to 42.8\% (+5.2). On full MMSI, it improves 2B/4B/8B backbones by 8.9/10.0/8.8 points; the corresponding MindCube gains are 19.9/21.5/20.8.
\begin{table*}[t]
\centering
\caption{\textbf{Robustness to controlled tool perturbations} on MindCube-Tiny (accuracy, \%). Coordinate noise is zero-mean Gaussian perturbation of $\pi^3$ outputs; mask noise randomly drops SAM3 foreground pixels. Hierarchical GRPO improves both clean accuracy and tolerance to corrupted evidence.}
\label{tab:main_robustness}
\resizebox{0.78\textwidth}{!}{%
\begin{tabular}{l c cc cc}
\toprule
\multirow{2}{*}{\textbf{Model setup}} & \multirow{2}{*}{\textbf{Clean}} & \multicolumn{2}{c}{\textbf{$\pi^3$ coordinate noise}} & \multicolumn{2}{c}{\textbf{SAM3 mask dropout}} \\
& & 0.2\,m & 0.4\,m & 15\% & 30\% \\
\midrule
Zero-shot tool use & 39.9 & 32.8 & 26.7 & 37.2 & 36.2 \\
w/o hierarchical GRPO & 51.2 & 44.3 & 40.3 & 49.2 & 49.0 \\
\textbf{Think-Capture (full)} & \textbf{54.4} & \textbf{50.7} & \textbf{46.2} & \textbf{53.8} & \textbf{52.1} \\
\bottomrule
\end{tabular}}
\end{table*}

\begin{table}[t]
\centering
\caption{\textbf{Backbone scaling} on MindCube-Tiny and full MMSI-Bench (accuracy, \%). Parentheses give absolute gains over each matched backbone.}
\label{tab:main_scaling}
\resizebox{\columnwidth}{!}{%
\begin{tabular}{l cc cc}
\toprule
\multirow{2}{*}{\textbf{Size}} & \multicolumn{2}{c}{\textbf{MindCube}} & \multicolumn{2}{c}{\textbf{MMSI full}} \\
& Base & Ours & Base & Ours \\
\midrule
Qwen3-VL-2B & 29.3 & 49.2 (+19.9) & 25.8 & 34.7 (+8.9) \\
Qwen3-VL-4B & 32.9 & 54.4 (+21.5) & 28.9 & 38.9 (+10.0) \\
Qwen3-VL-8B & 33.4 & 54.2 (+20.8) & 29.6 & 38.4 (+8.8) \\
\bottomrule
\end{tabular}}
\end{table}

\textbf{Robustness and representation controls.} With 0.4\,m coordinate noise, the full model retains 46.2\% accuracy versus 26.7\% for zero-shot tool use; with 30\% mask-pixel dropout, the scores are 52.1 and 36.2\%. A matched Stage-2 input with 2D boxes reaches only 36.9/31.8/37.9\% on MindCube/full MMSI/ViewSpatial, versus 54.4/38.9/42.8\% for symbolic 3D state. These controls support the value of learned 3D tool use relative to the tested visual-highlighting input; supplementary Sec.~C gives full perturbation and fallback results.

Table~\ref{tab:main_robustness} evaluates answer robustness as the quality of geometric evidence degrades. Under the strongest coordinate corruption, hierarchical rewards improve 40.3$\rightarrow$46.2\%, while the gap over zero-shot tool use is 19.5 points. Under 30\% mask dropout, the corresponding scores are 52.1, 49.0, and 36.2\%. The different degradation patterns are also informative: coordinate error changes metric relations, whereas pixel dropout often leaves enough target support for robust aggregation. In fewer than 1.6\% of MindCube-Tiny queries, the toolkit returns no valid target. In these cases, a deterministic fallback omits the empty symbolic state and completes the query using the original images.

\subsection{Generalization and Matched Controls}

Table~\ref{tab:main_scaling} applies the same acquisition, abstraction, and training recipe to 2B, 4B, and 8B Qwen3-VL backbones. MindCube gains remain within 19.9--21.5 points and full-MMSI gains within 8.8--10.0 points. The similar gains across model sizes suggest that explicit task-conditioned geometry complements backbone capacity. This pattern extends the benefit observed in the principal 4B comparison to both smaller and larger models.

\begin{table}[t]
\centering
\caption{\textbf{Matched visual-prompting control} (accuracy, \%). Both variants use the same Stage~1 target proposals and SAM3 masks.}
\label{tab:main_visual_control}
\resizebox{\columnwidth}{!}{%
\begin{tabular}{l ccc}
\toprule
\textbf{Stage-2 input} & \textbf{MindCube} & \textbf{MMSI full} & \textbf{ViewSpatial} \\
\midrule
2D boxes & 36.9 & 31.8 & 37.9 \\
\textbf{Symbolic 3D state} & \textbf{54.4} & \textbf{38.9} & \textbf{42.8} \\
\bottomrule
\end{tabular}}
\end{table}

\begin{figure*}[t]
    \centering
    \includegraphics[width=\textwidth]{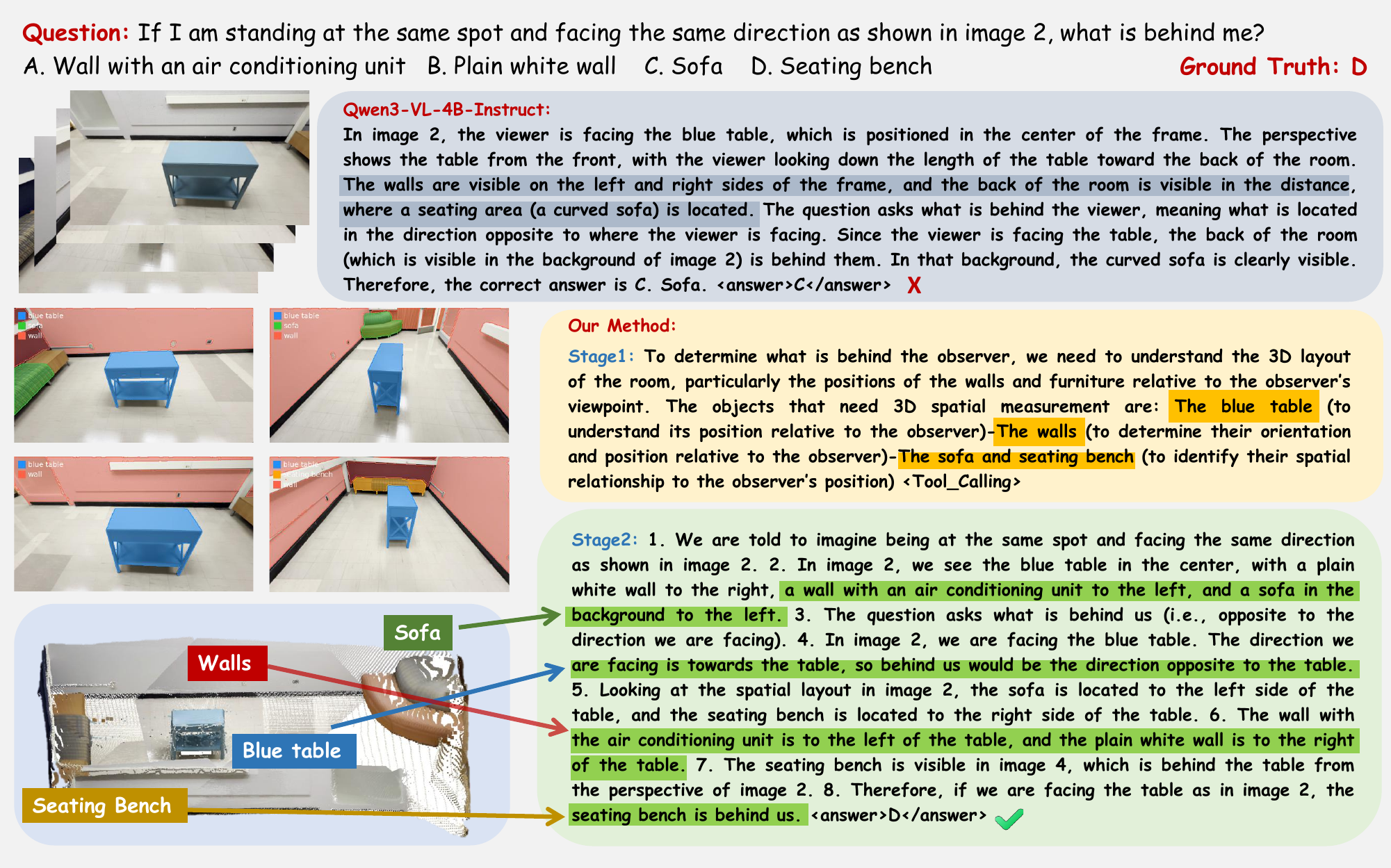}
    \caption{\textbf{Resolving egocentric spatial ambiguity on MindCube-Tiny.} The baseline equates the background visible beyond the blue table with the observer's rear and answers ``sofa.'' Think-Capture identifies the blue table as the facing anchor, relates the surrounding furniture in the common 3D frame, and correctly recovers the unseen seating bench behind the observer.}
    \label{fig:qualitative_1}
\end{figure*}

The matched control in Table~\ref{tab:main_visual_control} tests whether target localization or colored visual prompting alone explains the improvement. Replacing the symbolic state with 2D boxes derived from the same masks removes depth, metric scale, and camera pose while preserving which entities Stage~1 selected. It trails symbolic 3D state by 17.5 points on MindCube, 7.1 on full MMSI, and 4.9 on ViewSpatial. Under these tested input configurations, symbolic 3D state is more effective than 2D highlighting alone.

\begin{table}[t]
\centering
\caption{\textbf{Ablation study} on MindCube-Tiny. TC denotes Think-Capture; Tokens is the mean length of the 3D representation.}
\label{tab:ablation}
\resizebox{\columnwidth}{!}{
\begin{tabular}{lcccccccc}
\toprule
\textbf{Base Model} & \textbf{TC} & \textbf{GRPO} & \textbf{Format} & \textbf{Tokens} & Among & Around & Rotation & \textbf{Overall} \\
\midrule
Qwen3-VL-4B & & & - & - & 26.5 & 44.6 & 37.5 & 32.9 \\
Qwen3-VL-4B & & \checkmark & - & - & 28.9 & 41.2 & 36.4 & 33.3 \\
Qwen3-VL-4B & \checkmark & & - & 205 & 42.0 & 40.0 & 33.5 & 39.9 \\
Qwen3-VL-4B & \checkmark & \checkmark & Unified & 202 & 50.5 & 57.6 & 38.0 & 49.8 \\
\textbf{Qwen3-VL-4B} & \textbf{\checkmark} & \textbf{\checkmark} & \textbf{Template} & \textbf{100} & \textbf{57.6} & 54.1 & \textbf{45.0} & \textbf{54.4} \\
\bottomrule
\end{tabular}
}
\end{table}

\subsection{Ablation Study}
Table~\ref{tab:ablation} separates geometry acquisition, GRPO, and state formatting. GRPO alone gives only 32.9$\rightarrow$33.3\%, and zero-shot tool use reaches 39.9\%; combining the tools with GRPO reaches 54.4\%, showing that useful geometry and learned integration are complementary. Replacing the template with an unstructured coordinate dump reduces accuracy from 54.4 to 49.8\%. The template also uses 100 tokens per query, versus 202 for the unified format and 205 for raw tool output. Training curves in the supplement further show positive-sample accuracy stabilizing near 85\% while negative-sample accuracy rises from about 30\% to above 60\%, consistent with learning to use helpful geometry and reject misleading evidence.

\subsection{Qualitative Analysis}

Figure~\ref{fig:qualitative_1} asks what is behind an observer standing at the pose of image~2. This is not answered by locating what appears farthest away in that image: ``behind'' is the half-space opposite the camera's facing direction. The baseline follows the image-plane background and selects the sofa. Think-Capture first requests the blue table, walls, sofa, and seating bench. It uses the blue table as the visible facing anchor and the symbolic cross-view relations to place the seating bench on the opposite side of the observer. It therefore infers an object in the blind spot rather than copying a visible background object. The point cloud and arrows visualize the reconstructed spatial relations; Stage~2 takes the compact symbolic text and original frames as its inputs.

\begin{figure*}[t]
    \centering
    \includegraphics[width=1\textwidth]{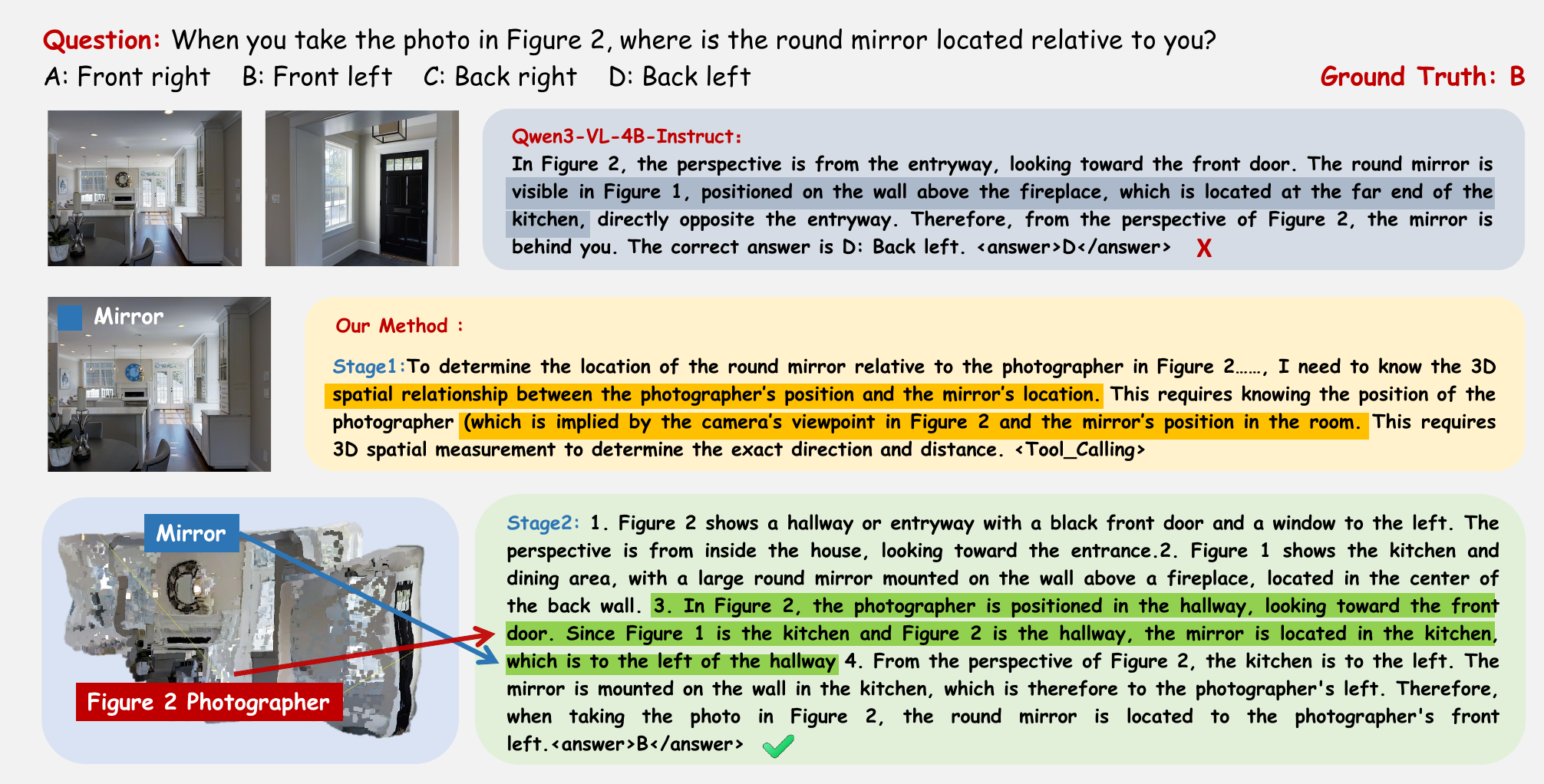}
    \caption{\textbf{Cross-view perspective taking on MMSI-Bench.} The baseline describes the mirror relative to the scene or to the wrong view and predicts ``back left.'' Think-Capture's response uses the photographer-relative description and gives the correct front-left answer.}
    \label{fig:qualitative_2}
\end{figure*}

The second complete case (Fig.~\ref{fig:qualitative_2}) isolates a different failure: the queried object and the reference camera are observed in different views. The baseline recognizes the mirror but narrates it relative to the house layout, not to the photographer in image~2, and consequently reverses front/back. Think-Capture explicitly captures the image~2 camera pose and the mirror geometry. The camera stream fixes the photographer's forward direction toward the front door; the object stream places the kitchen and mirror to the photographer's left. Transforming the mirror from the shared world frame into this egocentric frame yields \emph{front-left}. Together, the two cases show that the method addresses both unseen-space inference and cross-view reference-frame conversion.

\subsection{Joint Reasoning Dynamics}
The positive subset contains samples on which tool augmentation repairs a vision-only error, whereas the negative subset contains samples on which it overturns a correct visual answer. During GRPO training, positive-sample accuracy rises and stabilizes near 85\%, while negative-sample accuracy increases from about 30\% to above 60\%. Accuracy and mean reward eventually oscillate around a high plateau. Improvement on both screening subsets, together with the corruption results in Table~\ref{tab:main_robustness}, supports more robust integration of symbolic geometry and visual evidence. Together, the training curves and corruption tests characterize answer-level robustness across screening subsets and tool conditions.

\subsection{Real-World Deployment}
\label{subsec:deployment}
\begin{figure}[t]
    \centering
    \includegraphics[width=\linewidth]{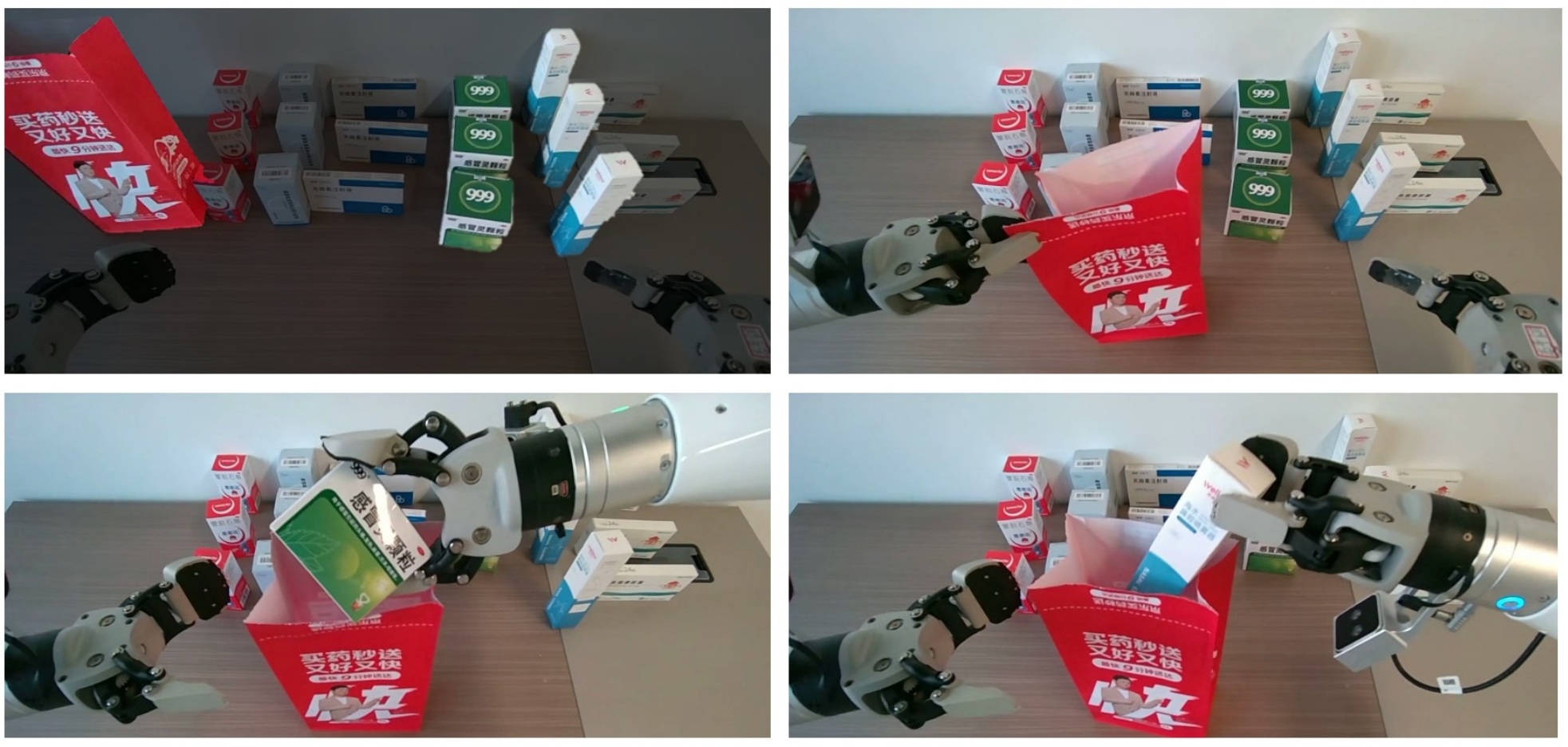}
    \caption{\textbf{Real-world deployment on Agibot G1.} Four snapshots show the initial observation, target approach, and medicine grasp. Think-Capture uses task-relevant symbolic geometry to supply a global waypoint to the high-frequency StarVLA motor policy.}
    \label{fig:robot_deployment}
\end{figure}

We deploy Think-Capture-4B as an asynchronous high-level spatial reasoner on an Agibot~G1 with the StarVLA platform~\cite{starvla2026} (Fig.~\ref{fig:robot_deployment}). At initialization, it captures task-relevant medicines and reference objects, resolves the target from symbolic 3D state, and supplies a low-frequency global waypoint; a separate StarVLA policy performs high-frequency arm control. The complete VLM+SAM3+$\pi^3$ stack peaks at 21.5\,GB on one RTX~4090. In the A100 benchmark, geometry takes 3.1\,s of the 21.8\,s pipeline (14\%); direct-answer decoding cuts total latency by over 40\% while retaining 53.2\% MindCube-Tiny accuracy. This architecture couples low-frequency spatial reasoning with high-frequency arm control; supplementary Sec.~D.2 describes the integration.

The timing breakdown is 6.8\,s for target planning, 3.1\,s for geometric capture, and 11.9\,s for Stage~2 reasoning. Think-Capture runs asynchronously: after receiving a waypoint, StarVLA continuously closes the visual-action loop while the high-level module is idle or processes the next request.

\textbf{Efficiency.} Stage~2 decoding is the largest latency term. Direct-answer decoding reduces end-to-end time by more than 40\% while retaining 53.2\% MindCube-Tiny accuracy, 1.2 points below full chain-of-thought decoding. The 21.5\,GB peak covers the VLM, SAM3, and $\pi^3$ resident together on one RTX~4090. Supplementary Sec.~D.1 details the measurement setup and runtime distribution.

\subsection{Geometry and Learned Integration}
Think-Capture combines pretrained perception backbones with target selection, symbolic state, and GRPO. Zero-shot tool augmentation reaches 39.9\%, while the complete configuration reaches 54.4\% (Table~\ref{tab:ablation}). The matched 2D-box comparison keeps target proposals and masks fixed and varies the Stage~2 representation (Table~\ref{tab:main_visual_control}). These results highlight the complementary roles of geometric acquisition, compact representation, and learned integration.

Table~\ref{tab:main_robustness} evaluates end-to-end answer accuracy under coordinate noise and mask dropout. The full configuration achieves higher accuracy across the tested corruption levels. Qualitative cases illustrate egocentric and cross-view reasoning, while gains across backbone sizes and on unseen MMSI and ViewSpatial benchmarks support transfer beyond the MindCube training set.

\textbf{Query-dependent computation.} Query-conditioned selection concentrates the symbolic state on the entities named by the question. Its computational benefit arises in segmentation, masked-point aggregation, and relation construction, where fewer targets yield fewer masks and object pairs. Camera poses remain available for viewpoint conversion. For a fixed set of retained frames, this separates the dense-estimation cost from the target-dependent cost of building the reasoning context, linking efficient acquisition to a compact, task-specific representation.

\subsection{Limitations and Scope}
Think-Capture depends on the observability and calibration available to its perception tools. A severely occluded target, systematic depth bias, or accumulated pose drift can make a symbolic relation wrong rather than merely uncertain. Hierarchical GRPO and the empty-target fallback reduce the effect of such failures but cannot recover geometry that is absent from every view. The current templates discretize continuous direction into linguistic bins suited to benchmark relations and waypoint selection. Fine manipulation additionally requires local shape and contact information at a higher spatial resolution.

In the robot deployment, Think-Capture selects task-relevant entities and generates a global waypoint, while the independently trained StarVLA policy handles fast visual feedback, collision-sensitive motion, and grasp execution. The 21.8\,s reasoning cycle operates asynchronously from StarVLA's high-frequency motor-control loop. Training uses disagreement-mined MindCube examples, with zero-shot evaluation on MMSI and ViewSpatial. Broader training scenes, uncertainty-aware symbolic states, and temporally persistent coordinates are promising directions for extending the framework to dynamic environments.

\section{Conclusion}
Think-Capture replaces passive dense fusion with query-conditioned geometry acquisition and a compact symbolic interface. GRPO trains the VLM to combine visual and 3D evidence, with improved robustness to noisy tool outputs in the tested settings. Consistent benchmark, perturbation, scaling, and real-robot results support active acquisition as a practical route to spatially grounded VLMs. The representation controls favor symbolic 3D state over the tested 2D-box input. The robot study demonstrates practical high-level integration. Future work will extend the interface to uncertainty-calibrated, temporally persistent geometry, enabling query-conditioned state refresh within the same hierarchical perception-and-control architecture.

\clearpage
\section*{Acknowledgments}
This work was supported in part by the Guangdong S\&T Programme (Grant No. \mbox{2024B0101030002}), the Basic Research Project of Hetao Shenzhen-HK ST Cooperation Zone (No. \mbox{HZQBKCZYZ-2021067}), the Shenzhen Outstanding Talents Training Fund (No. \mbox{202002}), and the NSFC (Grant Nos. \mbox{62293482}, \mbox{62573371}, \mbox{61931024}, and \mbox{12326610}). This work was also supported by the Guangdong Province Radio Science Data Center (Grant No. \mbox{2025B1212070001}), the Shenzhen General Program (No. \mbox{JCYJ20220530143600001}), the Guangdong Research Projects (Nos. \mbox{2017ZT07X152} and \mbox{2019CX01X104}), the Guangdong Provincial Key Laboratory of Future Networks of Intelligence (Grant No. \mbox{2022B1212010001}), the Shenzhen Key Laboratory of Big Data and Artificial Intelligence (Grant No. \mbox{SYSPG20241211173853027}), the National Key Research and Development Program of China (Nos. \mbox{2025YFF0515300} and \mbox{2025YFF0515304}), the Open Project Program of Key Laboratory of Tibetan Information Processing, Ministry of Education (Grant No. \mbox{QHSFCS2606}), the Shenzhen-Hong Kong Joint Funding (No. \mbox{SGDX20211123112401002}), and the Tencent \& Huawei Open Fund (Nos. \mbox{2024E0009} and \mbox{202301030019}).

{
  \small
  \bibliographystyle{ieeenat_fullname}
  \bibliography{arxiv_references}
}

\clearpage
\appendix
\setcounter{figure}{0}
\setcounter{table}{0}
\setcounter{equation}{0}
\renewcommand{\thefigure}{S\arabic{figure}}
\renewcommand{\thetable}{S\arabic{table}}
\renewcommand{\theequation}{S\arabic{equation}}
\renewcommand{\theHfigure}{supp.\arabic{figure}}
\renewcommand{\theHtable}{supp.\arabic{table}}
\renewcommand{\theHequation}{supp.\arabic{equation}}
\pdfbookmark[0]{Appendix: Supplementary Material}{thinkcapture.appendix}
\twocolumn[{\centering\Large\bfseries Supplementary Material\par\vspace{1em}}]
\section{Overview}
In this supplementary material, we provide additional details to support the main paper:
\begin{itemize}
    \item \textbf{Section~\ref{sec:implementation_details}} gives the SAM3+$\pi^3$ pipeline, symbolic abstraction, prompts, and training details.
    \item \textbf{Section~\ref{sec:revision_experiments}} reports controlled experiments on tool perturbations, empty-return behavior, backbone scaling, and the matched 2D-box baseline.
    \item \textbf{Section~\ref{sec:additional_results}} analyzes latency, memory, and real-robot deployment.
    \item \textbf{Sections~\ref{sec:limitations} and~\ref{sec:qualitative_analysis}} discuss limitations and provide additional qualitative cases.
\end{itemize}

\section{Implementation Details}
\label{sec:implementation_details}

\subsection{SAM3 + Pi3 Perception Pipeline}
\label{subsec:perception_pipeline}

The \texttt{AgentToolWrapper} combines text-prompted SAM3 segmentation with $\pi^3$ geometry estimation to expose an on-demand 3D interface to the VLM.

\subsubsection{Core Algorithm Setup}
The pipeline follows \textit{proposal $\rightarrow$ geometry $\rightarrow$ segmentation $\rightarrow$ fusion}. Stage~1 supplies the semantic targets $T$; only points selected by their masks enter the symbolic object state.

The core logic is implemented as shown in Algorithm~\ref{alg:pipeline}:

\begin{algorithm}[t]
\caption{Joint Semantic-Geometric Perception Pipeline}
\label{alg:pipeline}
\small 
\begin{algorithmic}[1]
\State \textbf{Input:} images $I=[I_1,\dots,I_N]$; targets $T$; thresholds $\tau_{conf},\tau_{dist}$
\State \textbf{Output:} $desc$: Structured 3D spatial analysis text

\Function{ExecutePipeline}{$I, T$}
    \State \textcolor{gray}{\# 1. Resource Management}
    \If{len($I$) $> 4$} $I \gets \text{uniform\_downsample}(I, target\_length=4)$ \EndIf

    \State \textcolor{gray}{\# 2. Geometric estimation ($\pi^3$)}
    \State $maps_{local}, poses \gets \pi^3.\text{forward}(I)$ 
    \State $memory \gets \text{SpatialMemoryManager()}$

    \State \textcolor{gray}{\# 3. Iterative Semantic-Geometric Fusion}
    \For{$k \gets 1$ \textbf{to} len($I$)}
        \For{$t$ \textbf{in} $T$}
            \State $masks,scores \gets \text{SAM3.segment}(I[k],t,\tau_{conf})$
            \For{$mask,score$ \textbf{in} \textbf{zip}($masks,scores$)}
                \State $points_{world} \gets poses[k] \cdot maps_{local}[k][mask]$
                \State $center,scale \gets \text{compute\_aabb}(points_{world})$
                \State $id \gets memory.\text{find\_nearest}(center,\tau_{dist})$
                \If{$id$ is found} $memory.\text{update}(id,points_{world},score)$
                \Else \ $memory.\text{add}(label=t,center=center,scale=scale)$
                \EndIf
            \EndFor
        \EndFor
    \EndFor

    \State \textcolor{gray}{\# 4. 3D Information Formatting}
    \State \Return $\text{format\_tool\_result}(memory.\text{get\_state}())$
\EndFunction
\end{algorithmic}
\end{algorithm}

\subsubsection{Implementation Specifics}
\begin{enumerate}
    \item \textbf{Resource-aware geometry.} The \texttt{Pi3Processor} uniformly downsamples sequences longer than four frames. Unconstrained long-sequence inference can exceed 57\,GB, whereas the four-frame setting keeps the complete deployed stack within 21.5\,GB.
    
    \item \textbf{Targeted segmentation.} Rather than running ``segment everything,'' the text-prompted \texttt{SAM3Wrapper} processes each target in $T$ and returns its instance masks and confidence scores. This reduces irrelevant masks and downstream point aggregation.
    
    \item \textbf{Spatial memory.} The manager maintains the cross-view object state.
    \begin{itemize}
        \item \textbf{Fusion:} Greedy matching merges an observation with an existing object when the centroid distance $\Delta d<\tau_{dist}$, with $\tau_{dist}=0.5$\,m.
        \item \textbf{State:} Each object stores its class, world position, axis-aligned size, and confidence.
    \end{itemize}
    
    \item \textbf{Coordinate convention and scale.} We use the Pi3X backend of $\pi^3$, which supports approximate metric-scale reconstruction.\footnote{\url{https://github.com/yyfz/Pi3}} The first camera defines the reference basis for direction words such as ``right'' and ``front,'' while the backend supplies the approximate metric scale. Distances, object sizes, fusion thresholds, and the 0.2\,m/0.4\,m perturbations use this common scale, with absolute accuracy depending on reconstruction quality. Robot execution additionally requires a mapping to the robot frame.
\end{enumerate}

\subsection{Symbolic 3D Abstraction: From Geometry to Semantics}
We convert masked point maps into the structured state used by Stage~2 through back-projection and symbolic formatting.

\paragraph{Object-level context.}
InstanceRefer~\cite{yuan2021instancerefer} grounds language in point clouds by combining instance attributes with instance--instance and instance--background context. Such contextual structure is also useful when a spatial question depends on a reference object rather than an isolated target. Think-Capture acquires the relevant geometry from images on demand and exposes object positions, extents, and relations through a compact symbolic interface. Richer object-level context could extend this interface to questions involving shape and contact.

\subsubsection{3D Back-Projection \& Attribute Extraction}
We compute 3D attributes by indexing the dense point maps with 2D semantic masks, discarding invalid points, and fitting axis-aligned bounding boxes (AABBs).

\begin{algorithm}[t]
\caption{3D Attribute Extraction}
\label{alg:extraction}
\small
\begin{algorithmic}[1]
\State \textbf{Input:} $mask_{2d}$, $map_{local}$, $pose_{cam}$
\State \textbf{Output:} $attr$: Object 3D attributes (center, scale)

\Function{ExtractAttributes}{}
    \State $points_{cam} \gets map_{local}[mask_{2d} > 0]$
    \If{len($points_{cam}$) == 0} \Return None \EndIf

    \State \textcolor{gray}{\# Transform to World Frame}
    \State $points_{world} \gets pose_{cam} \cdot points_{cam}$
    
    \State \textcolor{gray}{\# Compute AABB Stats}
    \State $center \gets \text{mean}(points_{world})$
    \State $scale \gets \text{max}(points_{world}) - \text{min}(points_{world})$

    \Return $\{ \text{"center"}: center, \text{"scale"}: scale \}$
\EndFunction
\end{algorithmic}
\end{algorithm}

\subsubsection{Spatial Reasoning \& Formatting}
The final step translates mathematical states into natural language. We categorize processing based on query type to filter irrelevant details.

\begin{algorithm}[t]
\caption{Symbolic 3D State Formatting}
\label{alg:formatting}
\small
\begin{algorithmic}[1]
\State \textbf{Input:} $memory$, $q\_type$
\State \textbf{Output:} $text$: Symbolic 3D description

\Function{FormatOutput}{}
    \State $lines \gets []$
    \State $objs \gets \text{GroupAndSelectTopK}(memory, K=2)$

    \If{$q\_type == \text{"cam-cam"}$}
        \For{$i, pose$ \textbf{in} enumerate($poses$)}
            \State $dir \gets \text{DescribeFacing}(pose)$
            \State $rel \gets \text{GetRelativePos}(pose_{ref}, pose)$
            \State $lines.\text{append}(\text{"Image " } + i + \text{": "} + dir + \text{", "} + rel)$
        \EndFor
    \Else
        \For{$obj$ \textbf{in} $objs$}
            \State $lines.\text{append}(obj.label + \text{: at } + obj.pos + \text{, size } + obj.scale)$
        \EndFor
        \For{$a, b$ \textbf{in} \text{pairs}($objs$)} \Comment{Compute Relations}
            \State $dir \gets \text{ProjectVector}(a.pos - b.pos, \text{frame}_{ref})$
            \State $lines.\text{append}(a.label + \text{ is } + dir + \text{ of } + b.label)$
        \EndFor
    \EndIf

    \Return $\text{join}(lines, \text{"\textbackslash n"})$
\EndFunction
\end{algorithmic}
\end{algorithm}

\textbf{Relative direction projection.} Spatial relations such as ``left'' and ``front'' are computed by projecting direction vectors onto the first frame's basis, aligning the description with the user's initial viewpoint.

\subsection{Prompt Templates}
\label{subsec:prompts}
Structured system prompts specify the planning (Stage~1) and reasoning (Stage~2) interfaces. The overview schematic uses ``simple/hard'' as shorthand for the positive/negative screening subsets, whose definitions follow the two screening models' answer correctness. Camera displacement is measured between views, and rotation is expressed in degrees. Stage~2 serializes its response as \texttt{<think>...</think><answer>...</answer>}, with a closing tag for each field. Explicit-state limitations also arise in memory-policy classification~\cite{chen2026explicit}.

\subsubsection{Stage 1: Intent Analysis}
\begin{promptbox}[System Prompt: Spatial Intent Analysis]
You are a spatial reasoning assistant with access to a 3D perception tool.
Output format:\\
<thinking>Analyze what spatial information is needed.</thinking>
<objects\_to\_segment>
List objects that need 3D measurement, one per line.
Example:
- car
- person
</objects\_to\_segment>

Question: \{question\}

Which objects need 3D spatial measurement to answer this question?
\end{promptbox}

\subsubsection{Stage 2: 3D-Augmented Reasoning}
\begin{promptbox}[System Prompt: 3D Reasoning]
You are a spatial reasoning assistant with 3D perception.
Use both the images and the 3D spatial data to answer.
Put the reasoning in <think>...</think> and combine visual cues with 3D information.
End with <answer>X</answer> where X is A, B, C, or D.

Question: \{question\}

3D Spatial Information:
\{formatted\_3d\}

Answer the question by combining the images above with the
3D spatial information. Think step by step, then give your
final answer.
\end{promptbox}

\subsection{Training Hyperparameters \& Details}
\label{subsec:hyperparameters}

We fine-tune Qwen3-VL-4B-Instruct with GRPO on four NVIDIA A100 (80\,GB) GPUs. Training stops at 1,600 steps to avoid overfitting the 3,032-example dataset.

\begin{table}[t]
\centering
\caption{Detailed hyperparameters for GRPO training.}
\label{tab:hyperparams}
\resizebox{\columnwidth}{!}{%
\begin{tabular}{ll}
\toprule
\textbf{Parameter} & \textbf{Value} \\
\midrule
Base Model & Qwen3-VL-4B-Instruct \\
Trainable Params & \begin{tabular}[t]{@{}l@{}}4,022M / 4,438M\\(vision encoder frozen)\end{tabular} \\
Optimizer & AdamW \\
Learning Rate & $5 \times 10^{-7}$, Cosine Schedule + 50-step Warmup \\
\midrule
\multicolumn{2}{l}{\textit{GRPO Configuration}} \\
Generations ($G$) & 8 per prompt \\
Clip Epsilon ($\epsilon$) & 0.2 \\
KL Penalty ($\beta$) & 0.04 \\
\midrule
\multicolumn{2}{l}{\textit{Reward Configuration}} \\
Accuracy Reward (Positive) & +1.0 \\
Accuracy Reward (Negative) & +2.0 (Anti-Regression) \\
Format Reward & +0.5 \\
\midrule
\multicolumn{2}{l}{\textit{Training \& Resources}} \\
Dataset Size & 3,032 samples (1,927 Pos / 1,105 Neg) \\
Effective Batch Size & 16 (Gradient Accumulation) \\
DeepSpeed Config & ZeRO-2, BF16 \\
Peak GPU Memory & 75.6 GB \\
Training Duration & \begin{tabular}[t]{@{}l@{}}$\sim$38 hours\\(Early stopped at 1,600 steps)\end{tabular} \\
\bottomrule
\end{tabular}}
\end{table}

\subsubsection{Training Dynamics}
\begin{figure*}[!tp]
    \centering
    \includegraphics[width=0.94\textwidth]{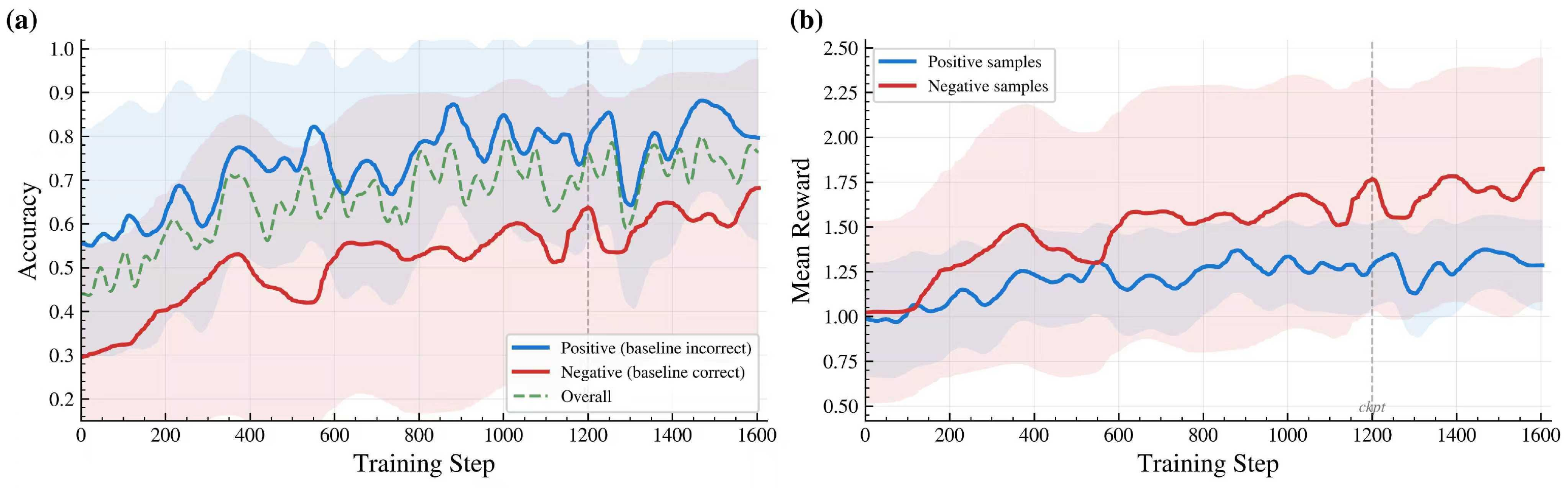}
    \caption{GRPO training dynamics: accuracy on positive/negative samples and mean reward.}
    \label{fig:supp_rl_dynamics}
\end{figure*}

\begin{figure}[!htbp]
    \centering
    \includegraphics[width=\columnwidth]{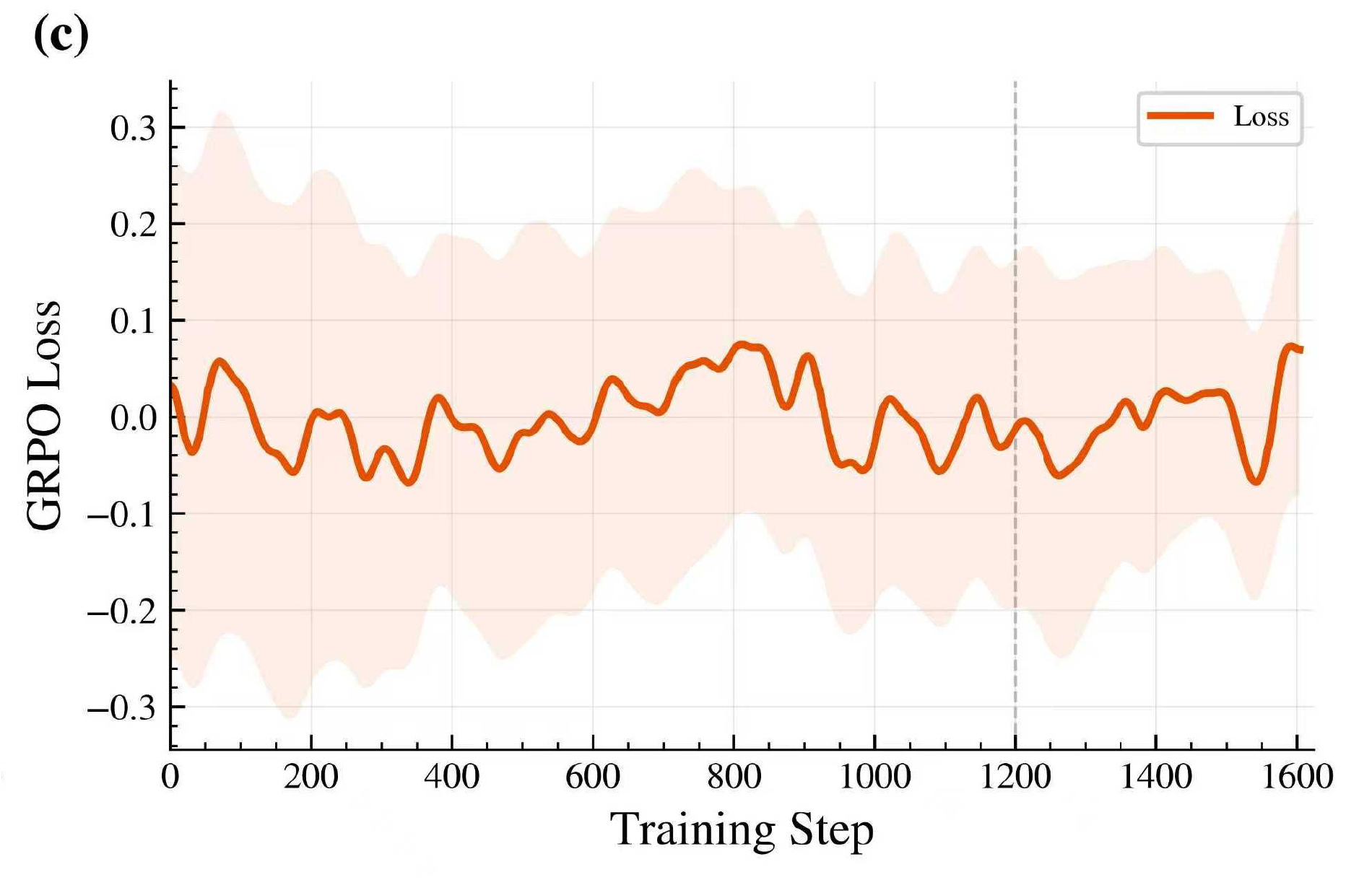}
    \caption{Training loss over the GRPO optimization trajectory.}
    \label{fig:training_loss}
\end{figure}

Figure~\ref{fig:supp_rl_dynamics} provides the full training dynamics behind the summary in the main paper. Figure~\ref{fig:training_loss} shows the corresponding optimization loss. Positive samples, for which 3D information is critical, stabilize near 85\% accuracy. Negative samples, for which geometry may be noisy or misleading, improve from roughly 30\% to above 60\%. The simultaneous improvement is consistent with better integration of visual and tool evidence across both screening subsets. Together with the perturbation experiments, these curves describe the evolution of answer-level robustness during training.

\FloatBarrier
\section{Additional Controlled Experiments}
\label{sec:revision_experiments}

These experiments examine robustness to imperfect tools, transfer across model scales and datasets, and the value of explicit 3D state beyond visual highlighting.

\subsection{Robustness to Tool Perturbations}
\label{subsec:tool_robustness}

We evaluate controlled degradation on MindCube-Tiny by adding zero-mean Gaussian noise to $\pi^3$ coordinates and randomly dropping SAM3 mask pixels. The zero-shot tool model degrades sharply, while the trained model remains substantially more accurate (Table~\ref{tab:perturbation}). The configuration without distinct positive/negative rewards has lower clean and perturbed accuracy. The comparison measures end-to-end answer robustness across training configurations and levels of tool corruption.

\begin{table}[t]
\centering
\caption{Accuracy (\%) under controlled tool perturbations on MindCube-Tiny.}
\label{tab:perturbation}
\resizebox{\columnwidth}{!}{%
\begin{tabular}{l c cc cc}
\toprule
\multirow{2}{*}{\textbf{Model setup}} & \multirow{2}{*}{\textbf{Clean}} & \multicolumn{2}{c}{\textbf{$\pi^3$ noise}} & \multicolumn{2}{c}{\textbf{SAM3 noise}} \\
& & 0.2\,m & 0.4\,m & 15\% & 30\% \\
\midrule
Zero-shot model & 39.9 & 32.8 & 26.7 & 37.2 & 36.2 \\
w/o hierarchical GRPO & 51.2 & 44.3 & 40.3 & 49.2 & 49.0 \\
\textbf{Ours (full)} & \textbf{54.4} & \textbf{50.7} & \textbf{46.2} & \textbf{53.8} & \textbf{52.1} \\
\bottomrule
\end{tabular}}
\end{table}

At the strongest coordinate perturbation, hierarchical GRPO improves accuracy from 40.3 to 46.2\%; at 30\% mask dropout, it improves 49.0 to 52.1\%. The tool returns no valid target in fewer than 1.6\% of MindCube-Tiny cases. A deterministic fallback omits the empty symbolic state and completes the query with an image-only response.

\subsection{Backbone Generalization}
\label{subsec:extended_generalization}

We apply the same training and inference pipeline to Qwen3-VL backbones at three scales. Gains are consistent across 2B, 4B, and 8B on MindCube-Tiny and full MMSI-Bench, indicating that the method is not specific to one capacity (Table~\ref{tab:backbone_scaling}). The main table reports the MMSI positional-relation subset; this table reports the full benchmark.

\begin{table}[t]
\centering
\caption{Backbone scaling on MindCube-Tiny and full MMSI-Bench. Parentheses show absolute gains over the corresponding base model.}
\label{tab:backbone_scaling}
\resizebox{\columnwidth}{!}{%
\begin{tabular}{l cc cc}
\toprule
\multirow{2}{*}{\textbf{Model size}} & \multicolumn{2}{c}{\textbf{MindCube-Tiny}} & \multicolumn{2}{c}{\textbf{MMSI full}} \\
& Base & Ours & Base & Ours \\
\midrule
Qwen3-VL-2B & 29.3 & 49.2 (+19.9) & 25.8 & 34.7 (+8.9) \\
Qwen3-VL-4B & 32.9 & 54.4 (+21.5) & 28.9 & 38.9 (+10.0) \\
Qwen3-VL-8B & 33.4 & 54.2 (+20.8) & 29.6 & 38.4 (+8.8) \\
\bottomrule
\end{tabular}}
\end{table}

\subsection{Explicit 3D State versus Visual Prompting}
\label{subsec:visual_prompting}

To test whether target highlighting alone explains the gains, we replace the Stage~2 symbolic state with frames annotated by 2D boxes derived from the same SAM3 masks. This controls for the target proposal and segmentation while removing depth, scale, and camera pose. It remains substantially below explicit 3D state on all three benchmarks (Table~\ref{tab:visual_prompting}), especially for egocentric and cross-view questions that require coordinate transformations.

\begin{table}[t]
\centering
\caption{Comparison with 2D-box visual prompting (accuracy, \%).}
\label{tab:visual_prompting}
\resizebox{\columnwidth}{!}{%
\begin{tabular}{l ccc}
\toprule
\textbf{Stage-2 input} & \textbf{MindCube} & \textbf{MMSI-Bench} & \textbf{ViewSpatial} \\
\midrule
2D boxes & 36.9 & 31.8 & 37.9 \\
\textbf{Symbolic 3D state} & \textbf{54.4} & \textbf{38.9} & \textbf{42.8} \\
\bottomrule
\end{tabular}}
\end{table}

The matched comparison favors symbolic 3D state over 2D-box prompting with the same selected targets and masks. Together with Table~\ref{tab:perturbation}, it characterizes the accuracy and robustness of the combined interface and training configuration. Target selection is shared by the two representation variants, and the reported outcomes are final-answer accuracies.

\section{Efficiency and Deployment}
\label{sec:additional_results}

\subsection{Efficiency \& Latency Analysis}
\label{subsec:latency}

We measure latency over all 1,050 MindCube-Tiny samples on one NVIDIA A100 (80\,GB).

\begin{table}[t]
\centering
\caption{Mean inference latency on MindCube-Tiny.}
\label{tab:latency}
\resizebox{\columnwidth}{!}{%
\begin{tabular}{llcl}
\toprule
\textbf{Stage} & \textbf{Component} & \textbf{Avg. Time} & \textbf{Description} \\
\midrule
Stage 1 & Planner \& Proposal & 6.8s & VLM analyzes query \& proposes targets \\
Tool & Geometry \& Perception & \textbf{3.1s} & Parallel $\pi^3$ + targeted SAM3 \\
Stage 2 & Spatial Reasoning & 11.9s & CoT reasoning \\
\midrule
\textbf{Total} & \textbf{End-to-End} & 21.8s & Median: 20.4s \\
\bottomrule
\end{tabular}}
\end{table}

\subsubsection{Latency Breakdown}
Tool execution accounts for 3.1\,s, or about 14\% of the 21.8\,s end-to-end time. Stage~1 planning (6.8\,s) and Stage~2 reasoning (11.9\,s) dominate because both generate text. Targeted segmentation and the four-frame budget keep perception comparatively small. When Stage~2 predicts the final answer without intermediate chain-of-thought tokens, end-to-end latency falls by more than 40\% while MindCube-Tiny accuracy remains 53.2\% (versus 54.4\% with full reasoning).

\subsubsection{Resource Consumption}
The complete VLM-4B, SAM3, and $\pi^3$ stack exhibits a peak GPU memory usage of \textbf{21.5 GB}. This indicates that the full 3D spatial-agent pipeline can be deployed on \textbf{consumer-grade hardware} such as an RTX~4090 with 24\,GB VRAM, lowering the deployment barrier compared with systems that require 40\,GB or 80\,GB accelerators.

\subsection{Real-World Deployment Details}
\label{subsec:deployment_details}
We integrate Think-Capture-4B with the StarVLA stack~\cite{starvla2026} on an Agibot~G1. At task initialization, Think-Capture receives the robot's visual observation, identifies the medicines and reference objects needed by the instruction, and invokes SAM3+$\pi^3$. The resulting target coordinates and relations are formatted exactly as in benchmark inference. Think-Capture resolves the target and emits a global spatial waypoint at low frequency; a separate StarVLA motor policy converts that waypoint into high-frequency arm commands. This separation keeps VLM generation outside the time-critical control loop. Similar planning-execution separation appears in robot control~\cite{zhao2026sim2real} and autonomous data collection~\cite{wang2026radar}. The four snapshots in the main-paper deployment figure show observation, approach, and grasp execution.

The deployment combines high-level semantic grounding with continuous visual-action feedback from StarVLA~\cite{starvla2026,zhang2026joyai,ra2026joyai}. Geometry is acquired at initialization, so moving targets and substantial scene changes require re-acquisition. With a 21.8\,s reasoning cycle, Think-Capture provides asynchronous waypoint updates; faster decoding and state refresh would extend this design to more reactive tasks.

\section{Limitations}
\label{sec:limitations}

Think-Capture has three main limitations:
\begin{itemize}
    \item \textbf{Foundation-model errors:} Although hierarchical GRPO and the empty-return fallback improve tolerance, severe $\pi^3$ reconstruction or SAM3 segmentation failures can still propagate to the answer.
    \item \textbf{Symbolic information loss:} Axis-aligned boxes and discrete relations omit fine shape, object orientation, and contact geometry needed for dexterous manipulation.
    \item \textbf{Latency and dynamics:} VLM generation dominates runtime, and the current robot integration captures geometry at initialization. Reactive deployment will require faster decoding and state refresh.
\end{itemize}

Broader training coverage, explicit uncertainty modeling, and efficient geometric state updates could improve robustness in dynamic environments~\cite{miao2026uncertainty,xu2025distributional,xu2024robust,xu2024uncertainty}.

\paragraph{Temporal extensions.}
Extending acquisition to changing scenes calls for tests that separate motion from appearance variation. Follow Your Pose~\cite{ma2024followpose} conditions character-video generation on body pose, while Follow-Your-Motion~\cite{ma2025followyourmotion} decouples spatial appearance and temporal motion for video motion transfer. These forms of control suggest ways to construct temporal reasoning cases with specified changes in pose or motion. For appearance variation, LiveLight~\cite{ma2026livelight} provides interactive 3D lighting control for streaming relighting with geometry-guided training, and LiveEdit~\cite{wang2026liveedit} performs causal streaming video editing with content preservation. Together, these directions offer tools for future evaluation of when geometric evidence should be refreshed under motion, lighting changes, or visual edits. Building such an evaluation would pair controlled sequences with verified spatial relations and time-dependent answer labels.

\onecolumn
\section{Additional Qualitative Analysis}
\label{sec:qualitative_analysis}

\subsection{Ambiguous Egocentric Relations}
Figure~\ref{fig:supp_case} provides a detailed comparison. The Qwen3-VL-4B baseline defaults to an image-plane relation (``back-right'') that is incorrect in the requested 3D frame. Think-Capture instead:
\begin{enumerate}
    \item \textbf{plans:} identifies the sofa and bag as the required targets;
    \item \textbf{perceives:} reconstructs them in a common coordinate frame; and
    \item \textbf{reasons:} interprets ``facing shelves'' in that frame to derive the correct egocentric relation.
\end{enumerate}

\begin{minipage}{\textwidth}
    \centering
    \includegraphics[width=0.94\textwidth]{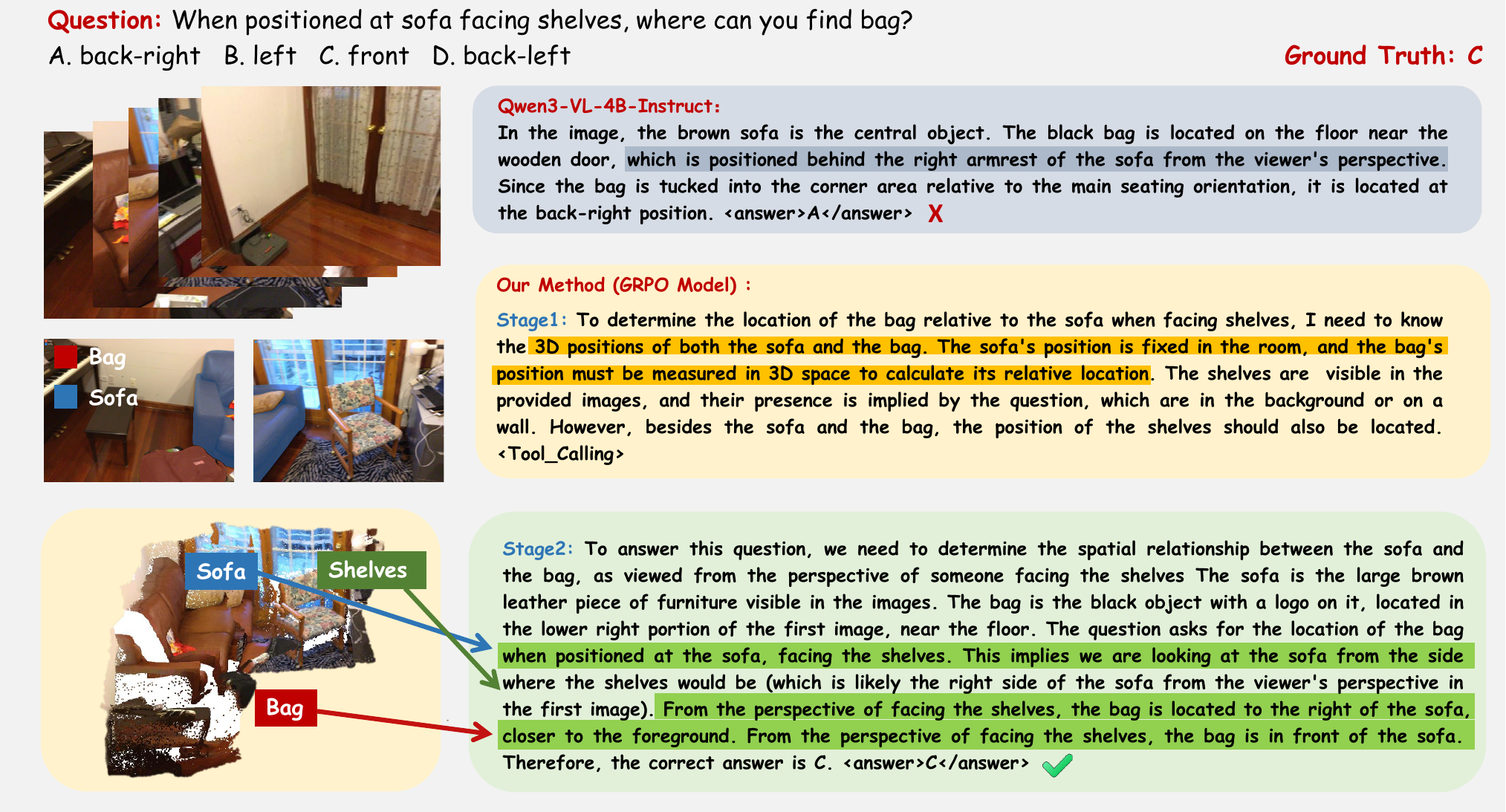}
    \captionof{figure}{\textbf{Ambiguous spatial reasoning.} The baseline predicts ``back-right'' from image-plane cues. Think-Capture selects the sofa and bag, grounds them with SAM3+$\pi^3$, and uses their symbolic state to infer the correct answer, ``front,'' in the requested frame.}
    \label{fig:supp_case}
\end{minipage}

\clearpage
\subsection{Egocentric and Cross-View Reference Frames}
Figures~\ref{fig:supp_qualitative_mindcube} and~\ref{fig:supp_qualitative_perspective} provide the complete MindCube-Tiny and MMSI-Bench examples discussed in the main paper.

\begin{minipage}{\textwidth}
    \centering
    \includegraphics[width=0.72\textwidth]{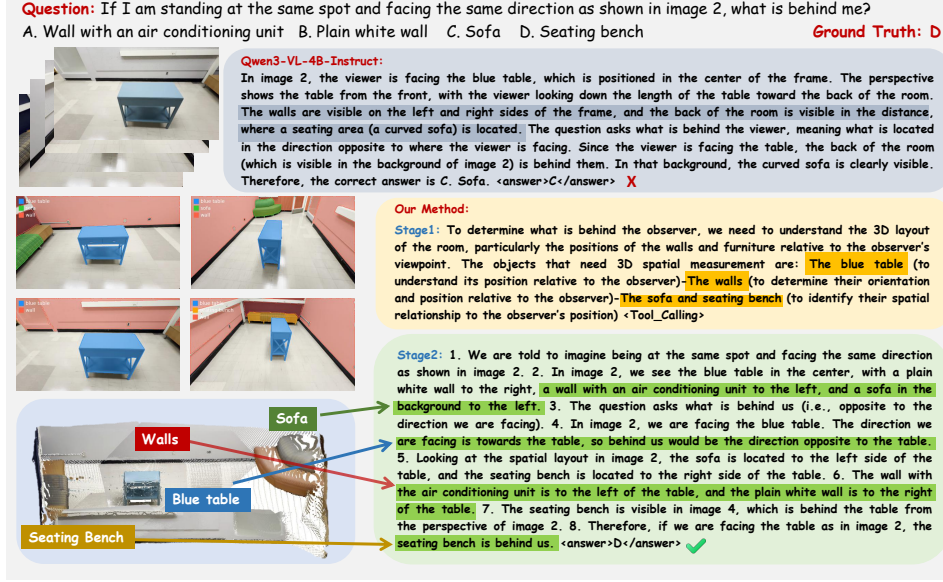}
    \captionof{figure}{\textbf{Spatial ambiguity on MindCube-Tiny.} The vision-only baseline confuses image-plane background with the observer's spatial rear. Think-Capture uses the Blue Table to establish facing direction and the Seating Bench relation to identify the unseen rear region.}
    \label{fig:supp_qualitative_mindcube}
\end{minipage}

\vspace{1em}
\begin{minipage}{\textwidth}
    \centering
    \includegraphics[width=0.94\textwidth]{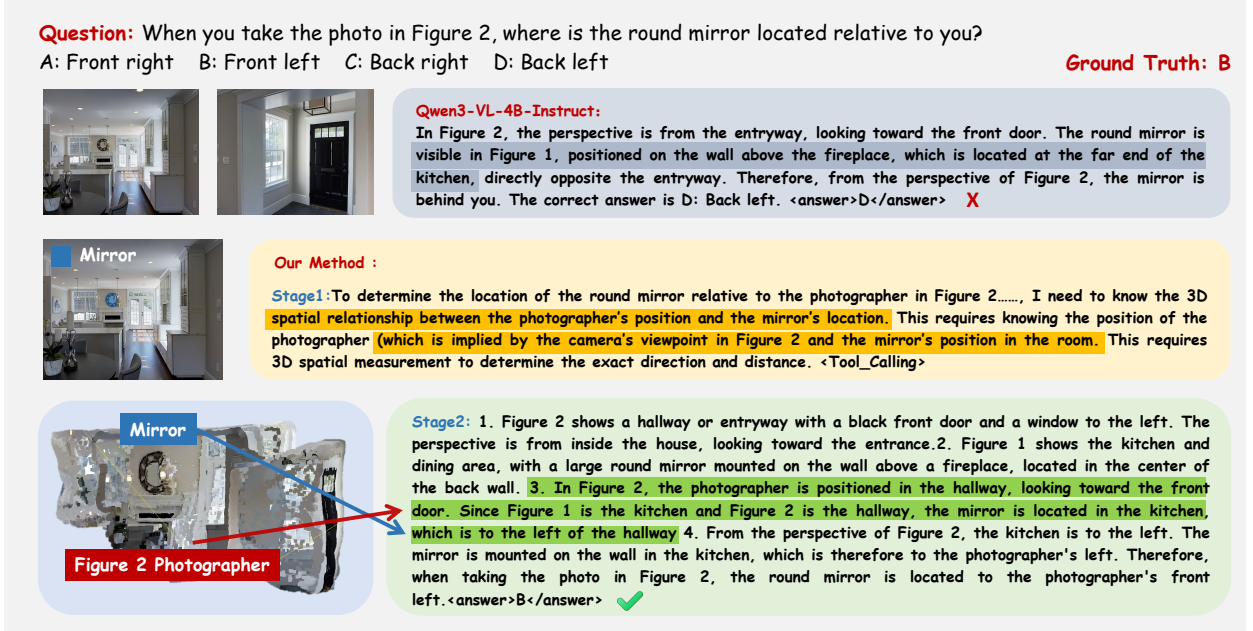}
    \captionof{figure}{Perspective-taking analysis on MMSI-Bench. Explicit camera pose and object geometry resolve the requested photographer-relative direction.}
    \label{fig:supp_qualitative_perspective}
\end{minipage}

\clearpage
\subsection{Additional Target-Selection Case}
Figure~\ref{fig:supp_case_one} illustrates the progression from target selection through geometry acquisition to the final answer.

\begin{minipage}{\textwidth}
    \centering
    \includegraphics[width=0.86\textwidth]{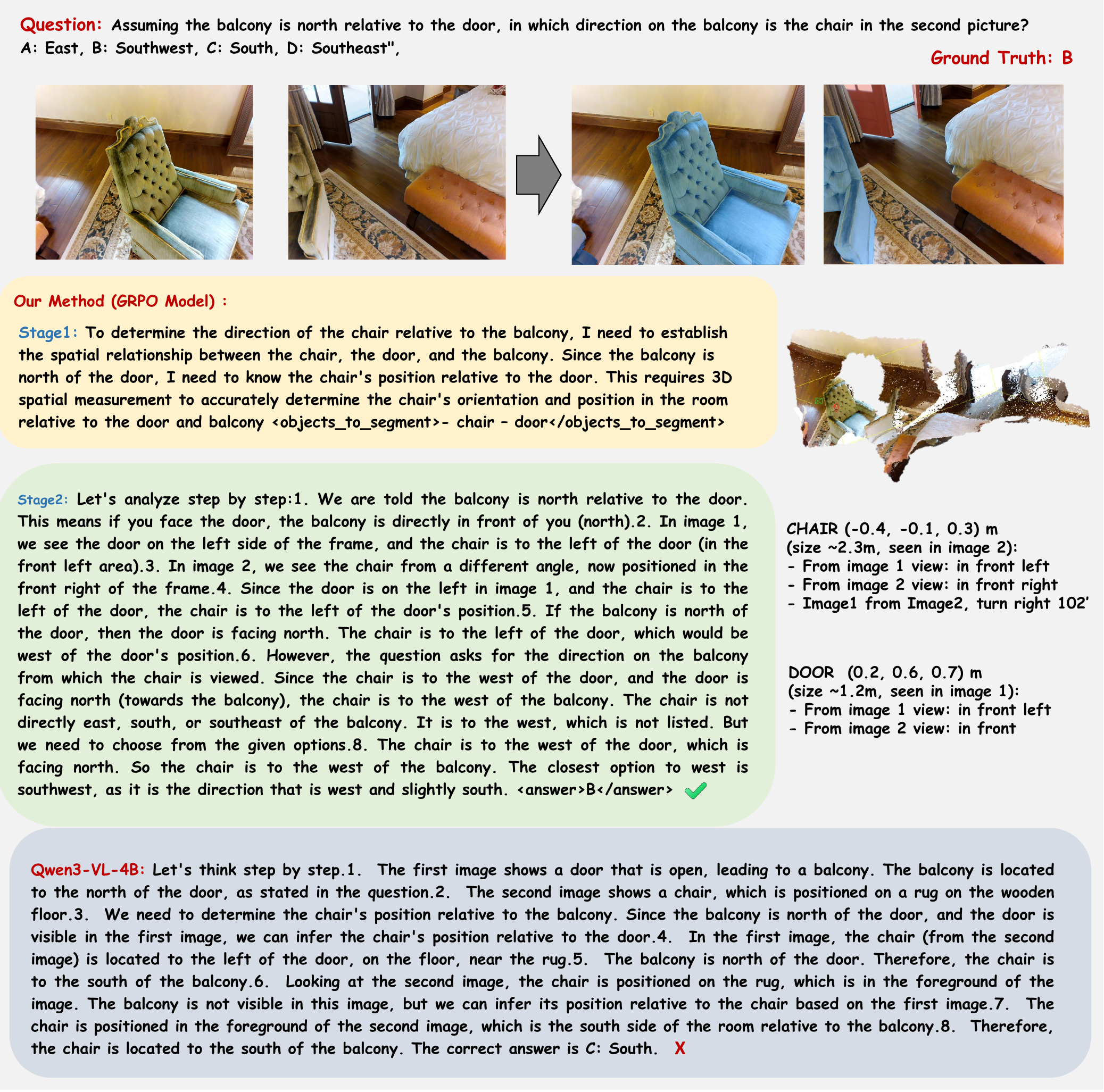}
    \captionof{figure}{Additional qualitative case illustrating target selection, reconstructed geometry, and the final spatial decision.}
    \label{fig:supp_case_one}
\end{minipage}

\twocolumn

\clearpage
\end{document}